\documentclass[10pt,conference,a4paper]{IEEEtran}
\usepackage{algorithm}
\usepackage[noend]{algpseudocode}
\usepackage{color,array}
\usepackage{xcolor}

\usepackage{times}
\usepackage{epsfig}
\usepackage{graphicx}
\usepackage{amsmath}
\usepackage{amssymb}
\usepackage{footnote}
\definecolor{bg}{rgb}{0.66, 0.66, 0.66}
\definecolor{vtl}{RGB}{189,198,255}

\definecolor{vrev}{RGB}{1,255,254}

\definecolor{vstrt}{RGB}{255,238,232}

\definecolor{talk}{RGB}{255,0,246}

\definecolor{opndr}{RGB}{107,104,130}

\usepackage{cite}
\ifCLASSINFOpdf
\else
\fi

\usepackage{hyperref}

\begin{document}
%
% paper title
% Titles are generally capitalized except for words such as a, an, and, as,
% at, but, by, for, in, nor, of, on, or, the, to and up, which are usually
% not capitalized unless they are the first or last word of the title.
% Linebreaks \\ can be used within to get better formatting as desired.
% Do not put math or special symbols in the title.
\title{\textit{TinyVIRAT}: Low-resolution Video Action Recognition}

% author names and affiliations
% use a multiple column layout for up to three different
% affiliations
\author{
\IEEEauthorblockN{Ugur Demir\IEEEauthorrefmark{1},Yogesh S Rawat\IEEEauthorrefmark{2} and Mubarak Shah\IEEEauthorrefmark{2}}
\IEEEauthorblockA{\textit{Center for Research in Computer Vision}\\\textit{University of Central Florida, Orlando, Florida, USA}\\
Email: \IEEEauthorrefmark{1}[ugur]@knights.ucf.edu, \IEEEauthorrefmark{2}[yogesh, shah]@crcv.ucf.edu}}

% conference papers do not typically use \thanks and this command
% is locked out in conference mode. If really needed, such as for
% the acknowledgment of grants, issue a \IEEEoverridecommandlockouts
% after \documentclass

% for over three affiliations, or if they all won't fit within the width
% of the page, use this alternative format:
%
%\author{\IEEEauthorblockN{Michael Shell\IEEEauthorrefmark{1},
%Homer Simpson\IEEEauthorrefmark{2},
%James Kirk\IEEEauthorrefmark{3},
%Montgomery Scott\IEEEauthorrefmark{3} and
%Eldon Tyrell\IEEEauthorrefmark{4}}
%\IEEEauthorblockA{\IEEEauthorrefmark{1}School of Electrical and Computer Engineering\\
%Georgia Institute of Technology,
%Atlanta, Georgia 30332--0250\\ Email: see http://www.michaelshell.org/contact.html}
%\IEEEauthorblockA{\IEEEauthorrefmark{2}Twentieth Century Fox, Springfield, USA\\
%Email: homer@thesimpsons.com}
%\IEEEauthorblockA{\IEEEauthorrefmark{3}Starfleet Academy, San Francisco, California 96678-2391\\
%Telephone: (800) 555--1212, Fax: (888) 555--1212}
%\IEEEauthorblockA{\IEEEauthorrefmark{4}Tyrell Inc., 123 Replicant Street, Los Angeles, California 90210--4321}}

% use for special paper notices
%\IEEEspecialpapernotice{(Invited Paper)}

% make the title area
\maketitle

% As a general rule, do not put math, special symbols or citations
% in the abstract
\begin{abstract}
The existing research in action recognition is mostly focused on \textit{high-quality} videos where the action is distinctly visible. In real-world surveillance environments, the actions in videos are captured at a wide range of resolutions. Most activities occur at a distance with a \textit{small resolution} and recognizing such activities is a challenging problem. In this work, we focus on recognizing tiny actions in videos. We introduce a benchmark dataset, \textit{TinyVIRAT}, which contains natural \textit{low-resolution} activities. The actions in \textit{TinyVIRAT} videos have multiple labels and they are extracted from surveillance videos which makes them realistic and more challenging. We propose a novel method for recognizing tiny actions in videos which utilizes a \textit{progressive generative} approach to improve the quality of low-resolution actions. The proposed method also consists of a weakly trained \textit{attention mechanism} which helps in focusing on the activity regions in the video. We perform extensive experiments to benchmark the proposed \textit{TinyVIRAT} dataset and observe that the proposed method significantly improves the action recognition performance over baselines. We also evaluate the proposed approach on synthetically resized action recognition datasets and achieve state-of-the-art results when compared with existing methods. 
\textit{The dataset and code is publicly available at \href{https://github.com/UgurDemir/Tiny-VIRAT}{https://github.com/UgurDemir/Tiny-VIRAT}.}
\end{abstract}

% no keywords

% For peer review papers, you can put extra information on the cover
% page as needed:
% \ifCLASSOPTIONpeerreview
% \begin{center} \bfseries EDICS Category: 3-BBND \end{center}
% \fi
%
% For peerreview papers, this IEEEtran command inserts a page break and
% creates the second title. It will be ignored for other modes.
\IEEEpeerreviewmaketitle

%%%%%%%%%%%%%%%%%%%%%%%%%%%%%%%%%%%%%%%%%%%%%%%
\section{Introduction}

% high resolution videos, talk about the resolution and all...

% tiny actions, existing works, synthetic dataset, use resizing ....

% we propose a realistic dataset...

% We also propose a solution, it works for existing synthetically created datasets as well...

% contributions...

Video action recognition has recently seen a good progress, which is mostly due to the availability of large-scale datasets and the success in deep learning. The availability of datasets, such as UCF-101 \cite{ucf101}, Kinetics \cite{kinetics}, Moments-in-time \cite{monfort2018moments}, AVA \cite{gu2018ava}, and Youtube-8M \cite{abu2016youtube}, has played an important role in this advancement. Apart from this, there are several novel architectures, such as C3D \cite{tran2015learning}, I3D \cite{i3d}, ResNet-50 \cite{resnet}, and TSN \cite{wang2016temporal}, which are proven to be effective for action recognition. However, the performance of these models relies on the quality of the action videos. The videos in all these datasets are of high quality and the action usually covers majority of the video frame. In real-world surveillance environments, the actions in videos are captured at a wide range of resolutions and may appear very tiny, therefore recognizing such actions becomes challenging at a very low-resolution. The existing action recognition models are not designed for low-resolution videos and their performance is still far from satisfactory when the action is not distinctly visible.

% Action recognition in real-world videos is challenging but crucial for video surveillance and monitoring applications. Recent advancements in convolutional neural networks (CNN) have improved action classification performance significantly. \MS{This has been mainly due to two reasons. First, large number of new action datasets e.g. Kinetics600 (get details from Swetha, she did a survey), EPIC Kitchen, etc with large number of classes and samples for each class. Second, significant progress in novel spatiotemporal CNN architectures give some examples here. However, their performance relies on the quality of the given input videos. Most datasets cover videos captured at high resolution by hand-held cameras by consumers..} In real-world scenarios, surveillance cameras are placed at distant places to cover a wide range of a scene. Actions that happen far away from the camera suffer from the lack of fine details and cause a performance drop. The performance of the existing action recognition methods on low-resolution (LR) action videos is still far away from being satisfactory.

%%%%%%%%%%%%%%%%%%%%%%%%%%%%%%%%%%%%%%%%%%%%%%%
\begin{figure*}[t!]
\begin{center}
\includegraphics[width=0.98\linewidth]{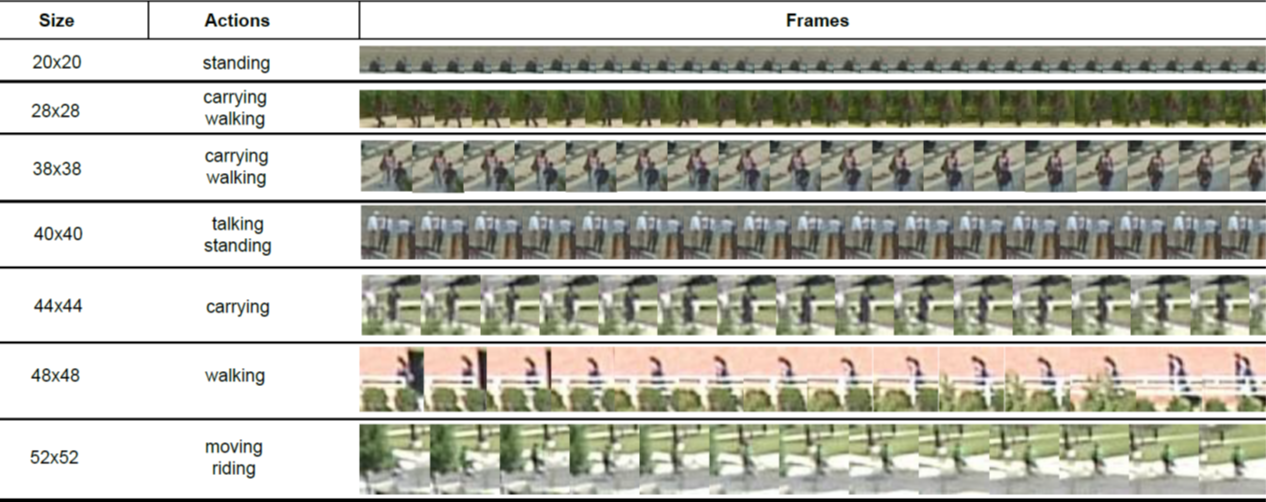}
\end{center}
    \caption{Some sample video frames for actions from \textit{TinyVIRAT} dataset. The dataset contain low-resolution videos with varying sizes. \textit{TinyVIRAT} is a multi-label dataset and each action video can have multiple action labels 
    %\MS{Also, replace last three examples with some other actions similar to other rows, where actions are very tiny.}
    }

  \label{fig:tiny_teaser}
\end{figure*}

%%%%%%%%%%%%%%%%%%%%%%%%%%%%%%%%%%%%%%%%%%%%%%%

In this work, our focus is on action recognition in low-resolution videos. The existing approaches addressing this issue, such as \cite{low_res_action_semi}, \cite{low_res_action_fully}, and \cite{zhang2019twostream}, perform their experiments on {\em artificially created datasets} where the high-resolution videos are down-scaled to a smaller resolution to create a low-resolution sample. However, re-scaling a high-resolution video to a lower- resolution does not reflect real world  low-resolution video quality. Real world low-resolution videos suffer from grain, camera sensor noise, and other factors, which are not not present in the down-scaled videos. 

% In the literature, the general approach for tiny action recognition problem is guiding action classifiers to extract similar features from both low-resolution (LR) and high-resolution (HR) videos  \cite{low_res_action_semi}, \cite{low_res_action_fully} or applying super-resolution to improve the video quality \cite{zhang2019twostream}. Although they show improvement on public benchmarks, the main issue is that the low-resolution videos are artificially created by down-scaling the videos from general-purpose action classification datasets. However, this approach does not reflect actual low-resolution video quality. Real low-resolution videos suffer from grain, camera sensor noise, and other factors.

To address this problem, we propose a new benchmark dataset, \textit{TinyVIRAT}, for low-resolution action recognition. The videos in \textit{TinyVIRAT} are realistic and extracted from surveillance videos of VIRAT dataset \cite{oh2011large}. This is a multi-label dataset with multiple actions per video clip which makes it even more challenging. The dataset has around 13K video samples from 26 different actions and all the videos are captured at 30fps. The length of the activities vary from sample to sample with an average length of around 3 seconds. It contains arbitrary sized low-resolution videos which ranged from 10x10 pixels to 128x128 pixels with an average of 70x70 pixels. The videos in the proposed dataset are naturally low resolution and they reflect real-life challenges. Some sample video frames from \textit{TinyVIRAT} are shown in Figure \ref{fig:tiny_teaser}. 
% \MS{Put some more details about the dataset; number of classes, number of videos, average resolution etc...}

% \MS{This and the next paragraphs are crucial, however they are very weak and low key. You should combine them and make them strong. They should basically capture  your figure 2. I do not know why you have figure 1. Or may be combine Figure 1 and 2.}To solve the tiny action recognition problem we \MS {we propose novel end to end approach, which has two three main parts progressive video super resolution, weakly supervised attention and action classification..... something like this.}introduce a new dense video super-resolution network (DVSR) and a progressive training procedure. The proposed super-resolution approach takes tiny actions and increases resolutions. It recovers important appearance and motion details by considering the spatio-temporal context of input frames. The progressive DVSR network starts from a low resolution (LR) and integrates new layers to model increasingly fine details as the training progresses. This operation recovers sharp details, which makes actions more distinguishable. 

We propose a novel end-to-end deep learning approach to address the problem of tiny action recognition. The proposed approach has three main components; video super-resolution, weakly supervised attention mechanism, and action classification. The video super-resolution network takes a low-resolution video and recovers important appearance and motion details using a Dense Video Super-Resolution network (DVSR), which is trained in a progressive manner. In this set up the foreground and background will have equal importance for the super-resolution task. However, foreground information has more discriminative information for action recognition. Therefore, to make DVSR action aware, we propose a novel attention mechanism which estimates a spatio-temporal map  indicating the importance of each pixel for the corresponding action. The attention map is trained in a weakly supervised setting which is guided by the action label of the video without requiring localization bounding boxes. The estimated spatio-temporal map is integrated with the synthesized high quality video to perform action recognition using a classifier.

% However, without additional guidance super-resolution training is action agnostic. The foreground and background have equal importance for the super-resolution task but foreground information has more discriminative information for the action recognition task. To make DVSR action aware, we added a new foreground attention branch that takes encoder features from DVSR and predicts a spatio-temporal weight map to indicate the importance of each pixel. The attention branch is trained by using only action labels. An action classifier is employed during DVSR training to guide the attention branch. This approach can be integrated with any differentiable action classifier model. We show that our approach performs better for the tiny action recognition task. We focus on action classification in this work; however, it is important to note that the proposed approach is a general method, which can be used for any other problem in the video domain such as anomaly detection, semantic segmentation, tracking, etc. 

In summary, this paper makes the following contributions:

\begin{itemize}
    \item We introduce a tiny action benchmark dataset which is the first dataset for this problem to the best of our knowledge.
    \item We propose a progressive video super-resolution based approach for tiny action recognition and demonstrate its effectiveness on \textit{TinyVIRAT} and artificial low-resolution action recognition benchmarks. 
    \item We also introduce a weakly supervised foreground attention mechanism that helps a super-resolution network to focus on important regions.
    \item We perform extensive experiments on the proposed \textit{TinyVIRAT} dataset and quantitatively demonstrate its challenging nature when compared with existing artificially created low-resolution benchmark datasets.
\end{itemize}

\section{Related work}

% \textbf{Action Recognition Dataset:} You want to talk about existing action recognition datasets and how the proposed one is different, it is tiny, multiple labels...

% \MS{This section also needs lots of work in re-organization. 
% Right now it is not coherent. For each paper you discuss you talk about how your paper is different, which is not a good idea. You first describe all classes of work and then at the end discuss what is missign and how your approach is different and better. For action classification subsection you should cover the work on datasets also, may be bring in first paragraph of Tiny Virat section here. Also, figure 2. Or leave it there.} 
\textbf{Action Classification:} After deep neural networks became popular for images, they have been successfully applied to the problem of video action recognition \cite{hara3dcnns}, \cite{tran2015learning}, \cite{i3d}. One of the popular deep network architecture C3D \cite{tran2015learning} showed that using 3D convolution is more suitable to extract spatio-temporal features for video action recognition. Recently, I3D architecture \cite{i3d} has shown favorable performance on standard benchmarks \cite{ucf101}, \cite{hmdb51}, \cite{kinetics} by employing Inception layers. In \cite{hara3dcnns}, deep ResNet \cite{resnet} architecture variants are investigated for the action recognition task.

For Low Resolution (LR) single image applications, several different approaches have been proposed, where domain adaptation, super-resolution or feature learning are employed to find better representations of LR images \cite{study_lrr}, \cite{tinyface_sr}. Previous work on this problem is generally motivated by privacy preservation \cite{towards_privacy},\cite{low_res_action_privacy}, \cite{low_res_action_multi_siamese}. In \cite{low_res_action_privacy},  a model is proposed which learns a set of different transformation that creates LR videos from the HR training set. It is claimed that action classifiers which are trained on the generated LR dataset learn better decision boundary. 
%\MS{I do not think, you want to bring in your method at this point. Just criticise this paper.} %We show that our SR based approach outperforms this inverse SR formulation by a large margin. 
In \cite{low_res_action_semi}, \cite{low_res_action_fully}, CNN based action classifiers are simultaneously trained with both LR and HR inputs by assuming that models learn similar representations. In \cite{zhang2019twostream}, the effect of super-resolution on the action recognition task is analyzed. They compared the existing image and video-based super-resolution networks, and proposed an optical flow guided training approach. However,  they only show their performance on artificially created low-resolution videos by downsampling UCF-101 and HMDB-51 dataset to 80x60, which is far from  natural tiny actions. %We introduce a new benchmark dataset to provide a more realistic problem setup for the low-resolution action recognition. 

\textbf{Super-Resolution:} One of the seminal CNN based single image SR method is proposed in \cite{srcnn}. After its success, several different CNN architectures have been introduced \cite{vdsr}, \cite{sr_laplace_pyramid}, \cite{edsr}. Although promising results are shown on single images, these methods cannot capture temporal information in videos if applied frame by frame. %In our SR model, we use 3D convolutions to synthesize temporally coherent videos. 
In some works, adversarial training has been utilized to obtain more realistic texture in images \cite{srgan}, \cite{enhancenet}, \cite{sr_dense}, \cite{multi_scale_res_sr}, \cite{abolghasemi2019pay}, nonetheless;  there is still no consensus about the best training scheme for SR models. Traditional pixel-wise reconstruction losses tend to produce smooth results, however, adversarial training introduces noise and artifacts \cite{Deng_2019_ICCV}. 
%We experimented with both reconstruction and adversarial losses and discovered that using L1 loss performs better for tiny action recognition. 

One of the most common strategies for Video SR is to incorporate optical flow for capturing motion information in order to synthesize a sequence of video frames \cite{frvsr}, \cite{vespcn}, \cite{f3}. In \cite{zhang2019twostream}, optical-flow is used to improve super-resolution network performance. The main drawback is that optical flow is computationally expensive, and if motion between frames is high, obtaining reliable estimates becomes difficult.
%In our approach, we put a weakly supervised foreground attention branch to highlight semantically meaningful regions instead of using optical-flow. 
In \cite{sr_temporal}, the 2D CNN network is used for frame SR and then temporal dependency weights are learned, which indicate how to merge input frames to synthesize the final frame. This method synthesizes one frame at a time. In \cite{peng2019progressive} the authors  proposed a 2D convolutional progressive network that incorporates short term inter-frame dependencies. There are some 2D convolution-based video SR approaches \cite{vsrnet}, \cite{sr_subpixel}, which only focus on short term temporal relations. In \cite{sr_recur}, recurrent CNN architecture is used to learn longer temporal dependencies for video SR. In some recent works \cite{vespcn}, \cite{sr_dynamic_up}, 3D convolution has been explored for effective video super-resolution. 

In contrast to these works, we propose a progressively growing architecture using 3D convolutions, where we start from a low-resolution and gradually increase the spatial and temporal resolutions. 
This progressive approach has been found effective in image generation \cite{prog_gan}, and we explore it for videos and demonstrate in this paper that it is  effective when we have a higher scaling factor. %\MS{This paragraph can be further strengthened to bring in limitations of both super resolution and action recognition related work and highlighting your end to end approach and emphasizing novelty. }

%%%%%%%%%%%%%%%%%%%%%%%%%%%%%%%%%%%%%%%%%%%%%%%

\begin{figure}[t!]
\begin{center}
\includegraphics[width=0.7\linewidth]{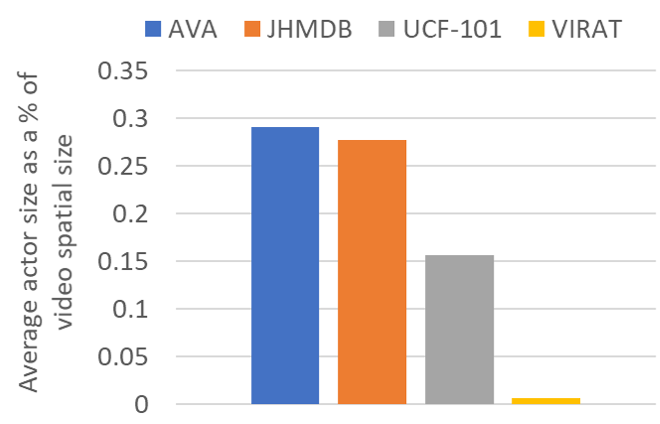}
\end{center}
%   \vspace*{-2cm}
  \caption{Average object size ratio comparison.}
  \label{fig:act_ratio}
\end{figure}

\begin{figure}[t!]
\begin{center}
\includegraphics[width=0.95\linewidth]{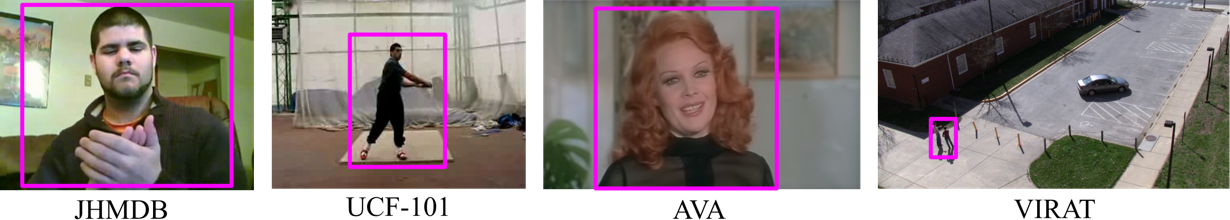}
\end{center}
%   \vspace*{-2cm}
  \caption{Comparison between avarage actor size and video resolution for different dataset. If the ratio is low enough, actions are naturally tiny.}
  \label{fig:ratio_samples}
\end{figure}

\section{TinyVIRAT Dataset}
Most of the existing action recognition datasets contain high resolution, actor centric videos \cite{Karpathy2014sports1m}, \cite{Xu2015A2D}, \cite{kinetics}, \cite{Charades}, \cite{youtube8m}, \cite{mpi2cooking}, \cite{avadataset}, \cite{caba2015activitynet}, \cite{THUMOS14}, \cite{Damen2018EPICKITCHENS}. For example, Kinetics \cite{kinetics}, Charades \cite{Charades}, Youtube-8M \cite{youtube8m} are collected from Youtube videos where actions cover most of the image  regions in every frame of a video. Using these videos to create low-resolution benchmark datasets does not reflect real world situation, and it is not appropriate as they generally contain larger actors.

In the real world, we encounter low-quality actions mostly in surveillance video clips where the camera placed in a distant place. Even though surveillance camera is capable of recording high-quality video, if an action happens far away from the camera, it will suffer from lack of details. Thus, surveillance videos are the perfect candidate for this problem. Figure \ref{fig:act_ratio} shows average actor size as a percentage of video spatial size, where most of the action recognition datasets have a significantly larger actor size. If the ratio is large, cropping actions will result similar spatial size with the original video.  In comparison, VIRAT dataset has naturally occurring tiny actors which is well suited for low-resolution action recognition task, as can be seen in Figure \ref{fig:ratio_samples}.

We introduced TinyVIRAT dataset which is based on VIRAT \cite{oh2011large} dataset for real-life tiny action recognition problems. VIRAT dataset is a natural candidate for low-resolution actions but it contains a large variety of different actor sizes and it is a very complex since actions can happen any time in any spatial position. To focus only on low-resolution action recognition problem, we crop small action clips from VIRAT videos.

In VIRAT dataset actors can perform multiple actions and temporally actions can start and end at different times. Before deciding which actions are tiny, we merged spatio-temporally overlapping actions and created multi-label action clips. We split these clips if the labels are changing temporally. This steps makes sure that created clips are trimmed. We selected clips that are spatially smaller than 128x128. Finally, long videos are split into smaller chunks and actions which do not have enough samples are removed from the dataset. We use the same train and validation split from the VIRAT dataset. 

\begin{figure}[t!]
\begin{center}
\includegraphics[width=0.95\linewidth]{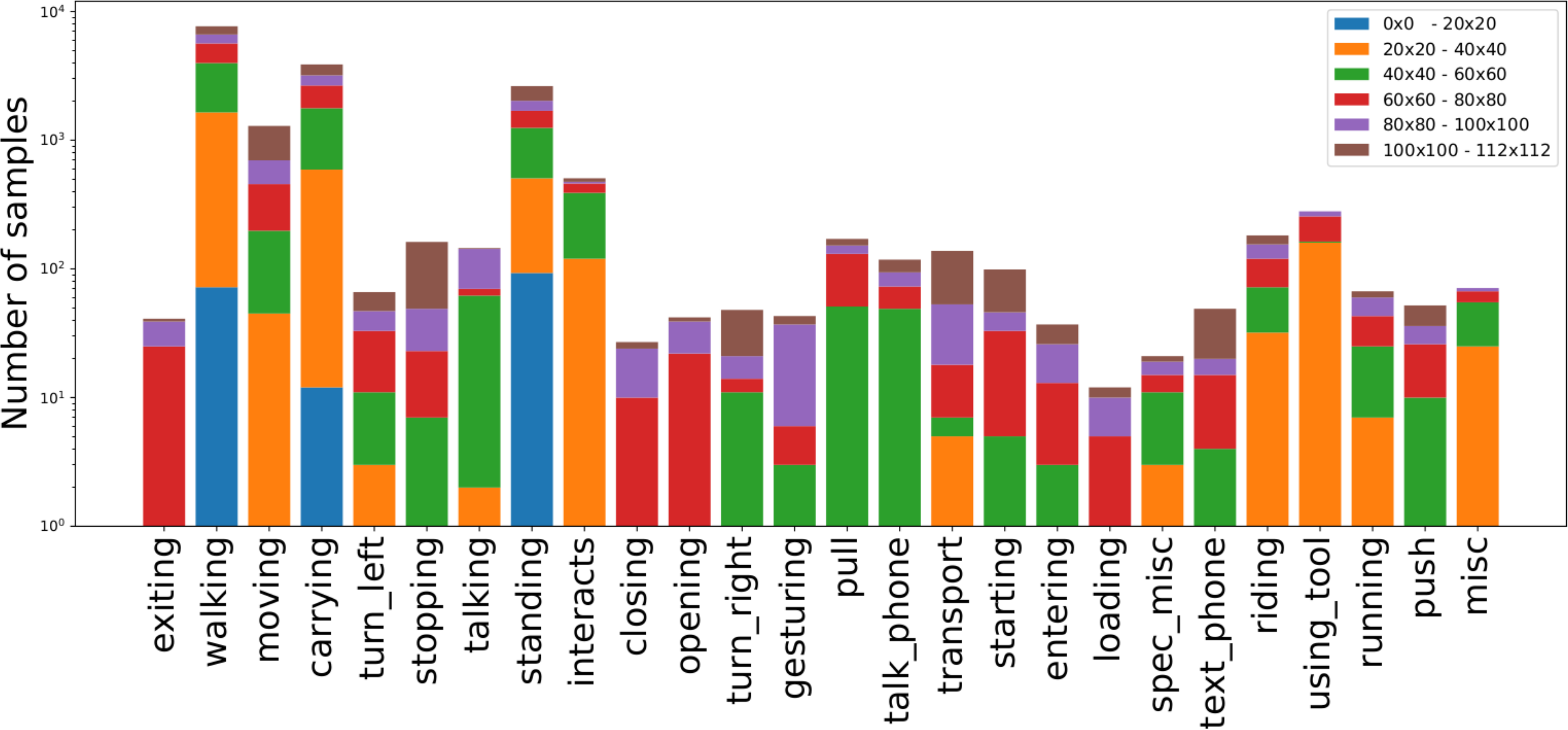}
\end{center}
%   \vspace*{-2cm}
  \caption{Number of samples per action labels and resolution. Numbers on y-axis are shown in log scale.}
  \label{fig:tiny_virat_stats}
\end{figure}

\textbf{TODO} TinyVIRAT has 7,663 training and 5,166 testing videos with 26 action labels. Table \ref{table:dataset_vs_stats} shows statistics from TinyVIRAT and several other datasets.  Figure \ref{fig:tiny_virat_stats} shows the number of samples per action class and the distribution of the videos by spatial size.

%%%%%%%%%%%%%%%%%%%%%%%%%%%%%%%%%%%%%%%%%%%%%%%%%%%%%%%%%%%%%%

\begin{table}[t!]
\begin{center}
\caption{Dataset statistics. ANF: Average number of frames, ML: Multi-label, NC: Number of classes, and NV: Number of vidoes.}
\label{table:dataset_vs_stats}
\begin{tabular}{l@{\hskip 0.25in}c@{\hskip 0.1in}c@{\hskip 0.1in}c@{\hskip 0.1in}c@{\hskip 0.1in}c}
\hline\noalign{\smallskip}
Dataset & Resolution & ANF & ML & NC & NV \\
\noalign{\smallskip}
\hline
\noalign{\smallskip}
UCF-101     & 320x240           & 186.50   & No  & 101 & 13320  \\
HMDB-51     & 320x240           & 94.49    & No  & 51  & 7000  \\
AVA         & 264x440 - 360x640 & 127081.66 & Yes & 80 & 272  \\
\hline
\textbf{TinyVIRAT}  & 10x10 - 128x128   & 93.93    & Yes & 26  & 12829  \\
\hline
\end{tabular}
\end{center}
\end{table}

% \vspace{-1cm}
%%%%%%%%%%%%%%%%%%%%%%%%%%%%%%%%%%%%%%%%%%%%%%%

%%%%%%%%%%%%%%%%%%%%%%%%%%%%%%%%%%%%%%%%%%%%%%%%%%%%%%%%%%%%%%
\begin{figure*}[t!]
\begin{center}
\includegraphics[width=0.98\linewidth]{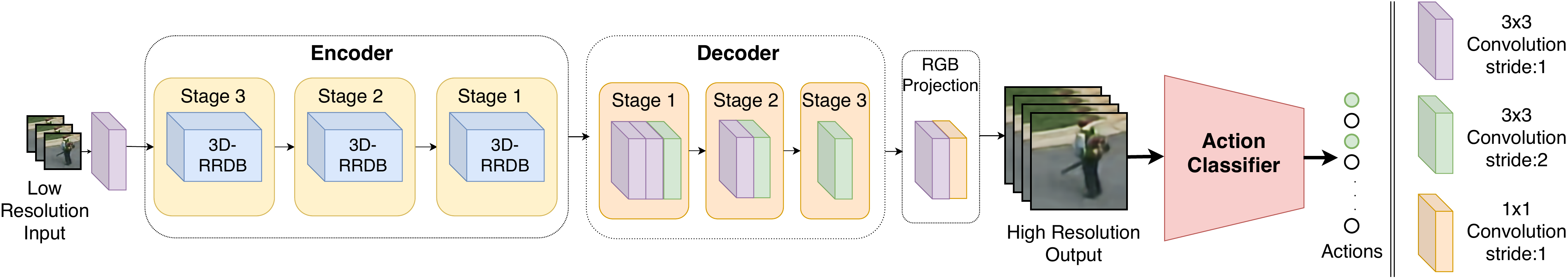}
\end{center}
    \caption{Overview of the progressive video generation and action classification approach. During the training process, we are introducing new blocks to Progressive DVSR network architecture at each stage. After video synthesis is completed, action classifier process the video to predict actions.  
    %\MS {There are no two blocks here a said in text: super resolution and classification. May be you should put the blocks.}
    %\MS{this is a better figure, I am not sure what is role of Figure 4; probably that is from old version of the paper, may be combine them.}
    }
  \label{fig:framework}
\end{figure*}

\section{Method}
The proposed method focuses on learning to enhance the quality of low-resolution videos to improve action classification performance. The action classifier network is trained with super-resolved videos instead of raw low-resolution video clips. Our approach consists of two main parts: (i) super-resolution and (ii) action classifier networks which can be seen in Figure \ref{fig:framework} 
%\MS{these blocks are not shown in the figure}. 
The first module, {\bf Super-resolution Network} (SR network),  is a novel deep convolutional neural network, which takes a low-resolution video clip and introduces sharp appearance and motion details to synthesize a high-resolution counterpart. The second module,  {\bf Action Classification Network},  takes the generated SR video clip and recognizes the action in the video. 
%The Super-resolution Network is first pre-trained to synthesize high-resolution videos. 
%We propose two different strategies for the pre-training of the SR network: (i) end-to-end and (ii) progressive \MS{why two? Reviewers will be confused}. Our super-resolution network has an additional foreground attention branch which guides our network to focus on important regions and learn features that are important for action recognition \MS{this should be emphasized more, since this is a new thing compared to other methods and it is making a difference. Right now this is mentioned on passing. First talk about what is the problem, and say you fix it with this branch} . Figure \ref{fig:sr_loc} shows weakly this supervised attention branch.

We propose progressive training strategy for the SR network which improves the reconstruction that helps action classification task. Improving texture quality leads us to get better performance but super-resolution network still does not know any task specific information. Focusing on the regions where the actions are happening is more beneficial for our main goal. Since, background of the videos does not have the importance as much as foreground for the action recognition task, guiding SR network is more crucial. We introduce a novel weakly-supervised foreground attention branch which guides our network to focus on important regions and learn features that are important for action recognition. Figure \ref{fig:sr_loc} shows weakly this supervised attention branch.

%\MS{this should be emphasized more, since this is a new thing compared to other methods and it is making a difference. Right now this is mentioned on passing. First talk about what is the problem, and say you fix it with this branch} . 

\subsection{Video Super-Resolution}
Video SR can be defined as finding sharp appearance and motion details from a low-resolution (LR) video to generate high-resolution (HR) video. We introduce a 3D convolution-based dense video SR (DVSR) network to solve this problem. The problem can be formulated as video-to-video translation, $\hat{V}_{HR} = G(V_{LR})$, where $V_{LR}$ is the low-resolution video clip, $\hat{V}_{HR}$ is the generated high-resolution video output and $G$ is the generator network, termed as DVSR network. 

%%%%%%%%%%%%%%%%%%%%%%%%%%%%%%%%%%%%%%%%%%%%%%%%%%%%%%%%%%%%%%

\subsubsection{Dense Video Super-Resolution (DVSR) Architecture} 
The proposed DVSR network consists of three main components, encoder, decoder, and a projection module. The encoder is responsible for feature extraction, the decoder part focuses on increasing the resolution of the features and the projection module generates the HR videos using those features. The use of residual blocks \cite{resnet} for image super-resolution tasks has been found effective in improving the image quality due to the similarities between the input and output \cite{vdsr}, \cite{esrgan}, \cite{enhancenet}. Motivated by this, we introduce 3D convolution-based residual-in-residual dense block (3D-RRDB). The 3D-RRDB module consists of a sequence of 3D-RDB modules integrated along with a residual skip connection (Figure \ref{fig:framework}). Each 3D-RDB module has densely connected five convolution layers \cite{densenet}. The input of 3D-RDB and the output is merged by a residual connection. 3D-RRDB contains three 3D-RDB modules and a skip connection from the first block to the last block. A detailed architecture of DVSR is shown in Figure \ref{fig:framework}.

The encoder takes low-resolution video frames and passes them through a 3D convolution layer and several 3D-RDB modules to extract important video features. 
%\UD{There is no pooling or sub-sampling operation during the encoding phase to preserve spatial size. Instead, we increase the depth of the network to determine distant spatial and temporal relations.}
The decoder part takes LR spatio-temporal features and projects them to HR space. Depending on the scale factor, the feature maps are up-scaled by linear interpolation followed by 3D convolutions. Instead of transposed convolution (fractionally strided convolution), using this strategy prevents checkerboard artifacts \cite{deconv}. Each up-scaling layer increases the spatial size by a factor of two. After completing spatial enhancement, obtained HR features are given to the projection module which consists of a sequence of 3x3x3 and 1x1x1 convolutions.

%%%%%%%%%%%%%%%%%%%%%%%%%%%%%%%%%%%%%%%%%%%%%%%%%%%%%%%%%%%%%%

\subsubsection{Progressive DVSR}
The proposed Progressive DVSR approach learns to increase the resolution in steps, with one scale at a time (Figure \ref{fig:framework}). We start with a shallow variant of DVSR architecture at the beginning of the training, which only increases the resolution by a factor of two. After it converges, we increase the depth of the encoder and decoder part of DVSR by adding new blocks, so that it can learn to generate features at a scale factor of four. This process is repeated until the desired resolution is obtained. This approach simplifies the problem by dividing it into multiple smaller steps.
% easier problem. We keep giving the same input to the network for all steps.

The encoder starts with a 3D convolution followed by a 3D-RRDB module. The decoder has one 3D upsampling along with 3D convolutions. Each step uses its projection layer. After progressing to the next step, previous projection layers are omitted. The network takes 16 RGB frames with 14x14 spatial resolution. In the first step, the network produces 4 frames of size 28x28. In the next step, a new 3D-RRDB module is added to the encoder and a new decoder module is added on top of the previous decoder. The network tries to synthesize 8 frames with a resolution of 56x56. We increase the temporal extent along with the spatial resolution. Therefore, the network learns to focus on necessary parts out of 16 frames. This progressive process continues until we have the desired resolution. 
%\MS {It will be good if you can try to express the progress process in a nice figure in the supplementary material.}

Each newly added layer causes a huge degradation in generated video quality. In order to make a smooth transition between progressive steps, we apply a fade-in operation to encoder and decoder separately. After each step, we keep using previous layer outputs along with new block outputs, since the new blocks produce noisy features. We decrease the effect of previous block outputs very smoothly until the new block is trained enough. The fade-in parameter $\alpha$ is set to 0 and gradually increased to 1 as the training progresses. 
%The details of the progressive step transition are shown in Figure \ref{fig:prog}.
In the encoder part, features from the previous layer are added to the new block output. In the decoder part, the output video clips are faded-in. Since each up-sampling block works on a different resolution, the previous projection layer's output is increased using linear interpolation. The final network architecture after completing the transition is similar to the end-to-end DVSR network.

%%%%%%%%%%%%%%%%%%%%%%%%%%%%%%%%%%%%%%%%%%%%%%%%%%%%%%%%%%%%%%

\begin{figure}[t]
\begin{center}
\includegraphics[width=0.98\linewidth]{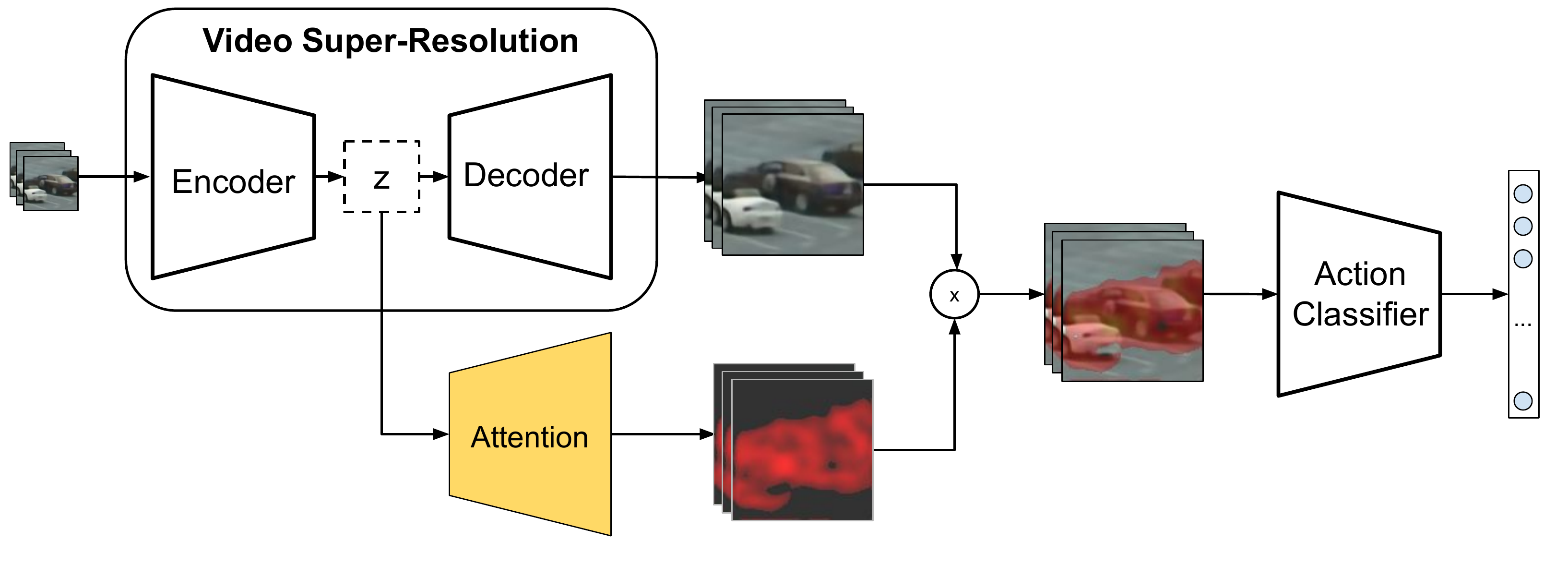}
\end{center}
    \caption{Overview of weakly-supervised foreground attention branch. It takes encoded features from the super-resolution network and predicts spatio-temporal importance weights for each pixel. During the training super-resolved video clip and the attention map is multiplied and given to the action classifier. The classifier guides the attention branch to highlight foreground regions. 
    %\MS{what is the weak supervision here?}
    }

  \label{fig:sr_loc}
\end{figure}

%%%%%%%%%%%%%%%%%%%%%%%%%%%%%%%%%%%%%%%%%%%%%%%%%%%%%%%%%%%%%%

\subsubsection{Foreground Attention}
Super-resolution networks generally focus on texture quality and reconstruction of the whole scene. They do not have knowledge about the foreground or background without guidance. Intuitively, we know that the foreground has much more importance for the action recognition task. To force our DVSR network to attach importance to prominent regions, we add a foreground attention branch to our DVSR network. It takes intermediate features from the encoder and predicts a spatio-temporal importance map for each frame. The predicted weights are used as a weight in DVSR training. 

The foreground attention branch is weakly supervised by using an action classifier network. During the training, generated HR video is masked with a predicted attention map and given to the action classifier. If the attention weights cluster around the foreground, action classifier should be able to classify the video. Otherwise, the attention branch will be penalized during the back-propagation. Figure \ref{fig:sr_loc} shows the training setup for the attention branch. 

\begin{figure}[!ht]
\begin{center}
\includegraphics[width=0.95\linewidth]{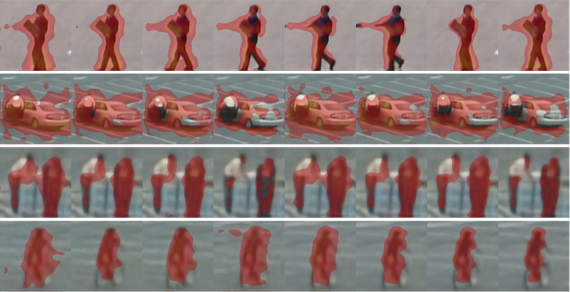}
\end{center}
    \caption{Examples of attention maps predictions by our weakly-supervised foreground attention branch. Attention maps are concentrated around the foreground actors. The predicted attention maps are used in super-resolution training to weight reconstruction loss. They successfully highlight the important regions for different cases; single actor, multiple actor, person object interaction at very low resolution settings.
    %\MS{Put more details here, what action each row represents. May be remove the first row, attention is not that good. Do not use "localization", use "attention" throughout. Otherwise reviewer may think you have annotations and your purpose is action localization. You have to say here or other place that your original purpose is not to localize action, but just recognize it. Attention  is  used to learn better representation.}
    }

  \label{fig:sample_loc}
\end{figure}

Training foreground attention branch by using only gradients from an action classification network can lead to a trivial solution that gives equally high importance to all of the pixels. If weights are equally important and high, all the pixels will be sent to the action classifier without filtering and there will be no feedback to the attention branch. To prevent that we use the L1 norm of the predicted attention map as a penalty so that large region predictions will be discouraged. Figure \ref{fig:sample_loc} demonstrates some of our weakly-supervised foreground predictions.

%%%%%%%%%%%%%%%%%%%%%%%%%%%%%%%%%%%%%%%%%%%%%%%%%%%%%%%%%%%%%%

\subsubsection{Super-resolution training}
The DVSR network takes LR input and synthesizes HR output. The difference between ground truth and generated video is used as a loss value to update network parameters. We employ a two-stage training strategy for the DVSR network. In the first stage, we pre-train our network progressive or end-to-end approach by using standard reconstruction loss. Afterward, we add the foreground attention branch and use the attention map to weight reconstruction loss.

\noindent \textbf{Reconstruction loss} is pixel-wise L1 distance between a ground truth video, $V_{HR}^i$, and the generated video, $\hat{V}_{HR}^i$. It is defined as:
\begin{equation}
\footnotesize
\mathcal{L}_{rec} = \frac{1}{N}\sum_{i=1}^{N}\frac{1}{CTHW}||\hat{V}_{HR}^i - V_{HR}^i||_1 \odot F_{att},
\end{equation}
where $N$ is number of samples in a batch, $C$, $T$, $H$, $W$ are channel size, number of input frame, height and width respectively. $F_{att}$ is the foreground prediction from the attention branch. For the first phase training $F_{att}$ is set to $1$.

\noindent \textbf{Foreground attention training}
The foreground attention branch is trained by an action classifier network. The predicted foreground attention map is applied to the reconstructed $\hat{V}_{HR}^i$ and it is used as an input to the action classifier. We use binary cross-entropy and cross-entropy for multi-label and single-label action classification tasks respectively. Also, we add L1 sparsity constraint to the predicted attention map and calculate the loss for attention branch by %\MS{may be you can bring in some of these points in the abstract and introduction to highlight novelty} 

\begin{equation}
\footnotesize
\mathcal{L}_{att} = - \frac{1}{N} \sum_{i=1}^{N} \sum_{c=1}^{K}y_c^i log(A(\hat{V}_{HR}^i \odot F_{att})) + \lambda_{att} ||F_{att}||_1,
\end{equation}

where $A$ is the action classifier, $\lambda_{att}$ is the coefficient for the regularization term and $y_c^i$ is the ground truth action labels.

The overall loss $\mathcal{L}$ is computed by combining both reconstruction and attention loss,
\begin{equation}
\footnotesize
\mathcal{L} = \lambda_{rec} \mathcal{L}_{rec} + \lambda_{att} \mathcal{L}_{att},
\label{eq:srloss_att_rec}
\end{equation}
such that each individual loss function is governed by a $\lambda$ coefficient.

%%%%%%%%%%%%%%%%%%%%%%%%%%%%%%%%%%%%%%%%%%%%%%%%%%%%%%%%%%%%%%

\subsection{Action classification}
%\MS{I do not think this is a good start of this section; do not just describe numerous experiments you did. Describe precisely what is your current method and why it is good.}We use several different architectures such as I3D \cite{i3d}, ResNet variants \cite{hara3dcnns} as our action classification network. The existing action classification networks assume that the input video has a certain resolution. Therefore, we observe a performance drop when they are used for low-resolution input videos, even after fine-tuning. We address this issue by using generative models to learn the distribution of high-resolution videos.

We use existing action classifier networks I3D \cite{i3d} and ResNet variants \cite{hara3dcnns} as our backbone structures. We combine our DVSR networks and classification network and use them as an end-to-end prediction model. Figure \ref{fig:framework} shows the final architecture. The super-resolution part takes the low-resolution videos and increase the spatial size while introducing new details. The synthesized high-resolution video is passed through the action classifier backbone to get final action prediction.

The main idea is that instead of using primitive interpolation methods on tiny action videos, our DVSR network is applied to improve LR video quality. Moreover, we utilize a weakly-supervised foreground attention prediction approach to highlight important features for the action classifiers. To achieve this, we use action classification as an auxiliary task for the super-resolution network during the joint training.    After that phase classifier is trained by using final DVSR network outputs without foreground attention branch. We use cross-entropy loss to train the action classifier network,
\begin{equation}
\footnotesize
\mathcal{L}_{act} = - \frac{1}{N} \sum_{i=1}^{N} \sum_{c=1}^{K}y_c^i log(A(G(V_{LR}^i))),
\end{equation}
where $ V_{LR}$ is Low-Resolution video,  $G$ is the super-resolution network, $A$ is the action classifier, $K$ is the number of classes and $y_c^i$ is the indicator function which is 1 if $c$  is equal to the given video label, otherwise, it will be zero.

%%%%%%%%%%%%%%%%%%%%%%%%%%%%%%%%%%%%%%%%%%%%%%%%%%%%%%%

%%%%%%%%%%%%%%%%%%%%%%%%%%%%%%%%%%%%%%%%%%%%%%%%%%%%%%%

\begin{table}[t!]
\begin{center}
\caption{Video Action Classification results for Tiny VIRAT dataset. % \MS{May be combine these three tables in one? or two? Otherwise, it appears you just want to fill the pages.}
}
\label{table:action:tiny_virat}
\begin{tabular}{lc}
\hline\noalign{\smallskip}
Method  & F1-Score \\
\noalign{\smallskip}
\hline
\noalign{\smallskip}
I3D                       & 28.73  \\ % 71a@0.2
I3D + Prog. DVSR          & 32.55  \\ % 92a@0.2
I3D + Prog. DVSR + Att.   &  34.49  \\ % 138a@0.2
\hline
ResNet-50                     & 29.08  \\
ResNet-50 + Prog. DVSR        & 29.81   \\
ResNet-50 + Prog. DVSR + Att. & 30.80   \\
\hline
WideResNet                        & 32.66  \\
WideResNet + Prog. DVSR           & 34.05  \\
WideResNet + Prog. DVSR + Attn.   & 35.07  \\
\hline
\end{tabular}
\end{center}
\end{table}

\section{Experiments}

\subsection{Datasets and Metrics}
We evaluate our approach on the proposed TinyVIRAT; TinyVIRAT has multi-label videos so performance is evaluated by F1 score. For super-resolution training, we crop the video clips from VIRAT following a similar strategy with TinyVIRAT but only allowing larger than 112x112 clips. To create low and high-resolution video pairs for super-resolution training, we down-scale and then up-scale the videos by using bicubic interpolation. 

In addition, we evaluate the performance of our system on two publicly available action dataset UCF-101 \cite{ucf101} and HMDB-51 \cite{hmdb51} in order to compare with existing methods.

%\MS{You should focus on TinyVIRAT; put one or two additional sentences here. Then you say in order to compare with other methods you also provide results on UCF-101 and HMDB. Otherwise, people are going to just focus on these two datasets instead of TinyVIRAT.}

%and two different public action classification datasets, UCF-101 \cite{ucf101} and HMDB-51 \cite{hmdb51}. UCF-101 has 13,320 videos from 101 different action classes. The original size of videos is 320x210 with a frame rate of 25 fps. HMDB-51 contains 7,000 videos for 51 categories with varying resolution and a frame rate of 25 fps. We use split 1 for both UCF-101 and HMDB-51 datasets in our evaluations. Since UCF-101 and HMDB-51 contain only single-label videos, we compute accuracy to evaluate the action classification performance. 
%For all of the datasets, low-resolution videos are obtained by artificially down-scaling and up-scaling again.

\subsection{Implementation Details}
DVSR is trained by setting $\lambda_{rec}$ to 1 and $\lambda_{att}$ to 0.5 in Equation \ref{eq:srloss_att_rec} during the joint training. Both DVSR and action classification networks are trained with Adam optimizer \cite{adam} and we use $\beta_1$ as 0.5 and $\beta_2$ as 0.9. Both the DVSR network and the action classifier are trained with a learning rate of 0.0002. The super-resolution network is trained without using any pre-trained model weights. The value of $\alpha$ for layer transition in a progressive approach is set to 0 initially and increased by 5e-3 after each iteration. This step size is set empirically and can be estimated based on the batch size and the number of epochs required for convergence.

For the action classification task, the I3D action classifier network is used with pre-trained weights which were obtained by training I3D on Charades dataset \cite{Charades}. ResNet architectures are pre-trained on Kinetics \cite{kinetics} dataset, we obtain the model parameters from \cite{hara3dcnns}.

%%%%%%%%%%%%%%%%%%%%%%%%%%%%%%%%%%%%%%%%%%%%%%%

%\begin{table}
%\begin{center}
%\caption{Video Action Classification results for Tiny VIRAT dataset.% \MS{May be combine these three %tables in one? or two? Otherwise, it appears you just want to fill the pages.}
%}
%\label{table:action:tiny_virat}
%\begin{tabular}{lcccccc}
%\hline\noalign{\smallskip}
%Method & Precision & Recall & F1-Score & Acc & Acc@top4 & Acc@top1 \\
%\noalign{\smallskip}
%\hline
%\noalign{\smallskip}
%I3D               & 18.89 & 13.58  & 42.38 & 51.72 & 90.73 & 87.63 \\
%I3D + DVSR        & 25.55 & 19.58  & 49.54 & 51.00 & 90.90 & 86.57 \\
%I3D + DVSR + Att. & 26.77 & 22.33  & 52.84 & 55.05 & 90.11 & 89.72 \\
%\hline
%ResNet-50               & 29.09 & 24.37 & 44.22 & 48.66 & 89.93 & 85.58 \\
%ResNet-50 + DVSR        & 24.74 & 19.17 & 47.80 & 49.63 & 90.73 & 86.59 \\
%ResNet-50 + DVSR + Att. & 25.12 & 19.27 & 49.71 & 49.75 & 90.22 & 86.59 \\
%\hline
%ResNet-101        & 26.59 & 20.88 & 45.59 & 50.91 & 90.82 & 87.17 \\
%ResNet-101 + DVSR & 30.12 & 25.78 & 45.59 & 51.06 & 89.61 & 88.41 \\
%\hline
%WideResNet          & 27.80 & 22.63 & 47.84 & 54.36 & 90.65 & 87.92 \\
%WideResNet + DVSR   & 30.94 & 25.27 & 50.31 & 54.53 & 89.76 & 88.56 \\
%\hline
%\end{tabular}
%\end{center}
%\end{table}

%%%%%%%%%%%%%%%%%%%%%%%%%%%%%%%%%%%%%%%%%%%%%%%

%\MS{I suggest you put put action classification sub-section before super resolution. You want reviewers to focus on action classification not super resolution, may be do not separate in two sub-section.}

\subsection{Quantitative Results}
We first train the action classifier networks standalone in order to set the baselines for TinyVIRAT. Spatial size of the low-resolution videos is resized to 112x112 by using bicubic interpolation. Then we apply our progressively trained DVSR network and foreground attention approaches to demonstrate improvement. For each experiment, the backbone action classifiers are initialized with the same pre-trained weights. 

Table \ref{table:action:tiny_virat} shows action classification performance baselines and our approach on TinyVIRAT dataset. Our DVSR network shows favorable improvement comparing to baseline network results. We experimented with different action classifier architectures to show that our DVSR is not biased to a certain network but improves the results consistently for all the models. After introducing our weakly-supervised foreground attention approach we improve the baseline scores by a large margin.

%We first train the action classification network with the HR ground truth videos to set upper bound for the recognition performance. As a baseline, we train the same network with LR videos by increasing their spatial size using bicubic interpolation to a set the baseline for TinyVIRAT. Table \ref{table:action:tiny_virat} shows action classification performance baselines and our approach on TinyVIRAT dataset.

We also compare our method with previous work \cite{low_res_action_privacy} \cite{vespcn} and \cite{zhang2019twostream} on public datasets. For a fair comparison, we compare our method with other non-optical flow based methods. Table \ref{table:action:hmdb} and Table \ref{table:action:ucf} show the results on HMDB-51 and UCF-101 respectively.

\begin{table}[t!]
    
      \caption{Video Action Classification comparison for HMDB-51 dataset.}
      \label{table:action:hmdb}
      \centering
        \begin{tabular}{llc}
            \hline
            \textbf{Method} & \textbf{Input} & \textbf{Accuracy \%} \\
            \hline
            %I3D (64 frames) \cite{i3d} & 224x224 & 49.80\% \\ 
            I3D & 112x112 & 52.61 \\ % 121a
            \hline
            SoSR \cite{zhang2019twostream} & 80x60 & 54.77 \\
            \hline
            %Semi-Coupled \cite{frvsr} & 12x16 (rgb+flow) & 29.20\% \\
            %Siamese \cite{low_res_action_multi_siamese} & 12x16 (rgb+flow) & 39.70\% \\
            %F. Coupled \cite{vespcn} & 12x16 (rgb+flow)  & \textcolor{red}{44.96\%} \\
            %\hline
            Bicubic - I3D & 14x14 & 10.59 \\ %123a
            Privacy-Preserv \cite{low_res_action_privacy} & 12x16  & 28.68 \\
            F. Coupled \cite{vespcn} & 12x16  & 39.15 \\
            %\hline
            DVSR & 14x14 & 41.24 \\ % 136sa
            {\bf Prog. DVSR} & {\bf 14x14} & \textbf{41.63} \\ 
            \hline
            Bicubic - I3D & 28x28 & 46.97 \\ % 124a
            Privacy-Preserv \cite{low_res_action_privacy} & 24x32 & 32.15 \\
            %\hline
            DVSR & 28x28 & 53.66 \\ % 116sa
            {\bf Prog. DVSR} & {\bf 28x28} & \textbf{55.95} \\ 
            \hline
        \end{tabular}
    
\end{table}

\begin{table}[t!]
        \centering
        \caption{Video Action Classification comparison for UCF-101 dataset.}
        \label{table:action:ucf}
        \begin{tabular}{llc}
            \hline
            \textbf{Method} & \textbf{Input} & \textbf{Accuracy \%} \\
            \hline
            %I3D (64 frames) \cite{i3d} & 224x224 & 84.50\% \\
            I3D  & 112x112 & 84.72 \\
            \hline 
            SoSR \cite{zhang2019twostream} & 80x60 & 83.92\\
            \hline
            Bicubic - I3D & 14x14 & 14.14 \\ % 122a
            DVSR & 14x14 & 68.17 \\
            {\bf Prog. DVSR} & {\bf 14x14} & \textbf{70.55} \\
            \hline
            Bicubic - I3D & 28x28 & 66.72 \\ % 90a      70.56\% 
            DVSR  & 28x28 & 82.37\% \\ % 99sa
            {\bf Prog. DVSR}  & {\bf 28x28} & \textbf{82.87} \\ % 
            \hline
            \end{tabular}
\end{table}

%%%%%%%%%%%%%%%%%%%%%%%%%%%%%%%%%%%%%%%%%%%%%%%

\subsection{Ablation study}
We have already shown that using super-resolution improves the action classification results. We experiment with different strategies to study the variation in the performance of the action classification task. Table \ref{table:ablation} shows the performance for different training strategies. For each experiment, the action classifier part uses the same action classification loss and for SR experiments we use DVSR network. We only change the generator training strategy while training.

\subsubsection{Non-progressive Super-Resolution Training}
We show that using the standard end-to-end super-resolution approach, action classifier performance can be boosted. In Table \ref{table:ablation}, the first two rows show the effect of using a standard super-resolution approach.% The standard training procedure helps us to get nearly 5\% improvement.

\subsubsection{Progressive Training}
Progressive training strategy simplifies the optimization process for the super-resolution problem. Instead of learning the mapping between low-resolution to high-resolution, it breaks down the process into smaller tasks. As we expected the proposed progressive DVSR model provides a significant improvement over the baseline approach. It also improves the standard super-resolution performance as we can see in the third row of Table \ref{table:ablation}. 

\subsubsection{Foreground Attention Branch}
Using weakly-supervised foreground attention branch while super-resolution network training gives us the best result. It guides DVSR network to focus on foreground regions that have more meaningful information for the action recognition task. Figure \ref{fig:sample_loc} shows qualitative results for super-resolution and attention maps. From the visual results, we can see that when the quality of the video is low, distinguishing the background and foreground becomes more difficult but our attention prediction successfully concentrates around the actors. The improvement we get from attention guidance is orthogonal to progressive training strategy, and using both of them together lead us the best performance as we can see in the last row of Table \ref{table:ablation}.

%%%%%%%%%%%%%%%%%%%%%%%%%%%%%%%%%%%%%%%%%%%%%%%

\section{Conclusion}
We introduce a new tiny action recognition benchmark dataset \textit{TinyVIRAT} which consists of natural low-resolution videos. We propose a novel tiny video action classification framework which incorporates progressively growing video super-resolution network to improve tiny action recognition performance. We utilize a 3D convolution-based dense residual network and weakly-supervised foreground attention branch which helps in learning effective appearance and motion features from low-resolution videos. We also demonstrate that the enhanced videos using the progressive DVSR network learn important appearance and motion features which is beneficial for action recognition. We perform experiments on two artificial benchmark datasets and demonstrate that the proposed approach leads to a better action recognition performance on both artificially created and natural low-resolution videos. Our super-resolution based tiny action classification framework can be integrated into any video analysis pipeline without much effort for other problems such as semantic segmentation, object localization, and object tracking.

\begin{table}[t!]
\begin{center}
\caption{Ablation results. Our novel progressive training strategy and weakly-supervised attention approach improves the performance of standard action classifiers. We use I3D architecture for these experiments.}
\label{table:ablation}
\begin{tabular}{lc}
\hline\noalign{\smallskip}
Method  & F1-Score \\
\noalign{\smallskip}
\hline
\noalign{\smallskip}
w/o DVSR                     & 28.73 \\ % 71a@0.2
DVSR                          & 30.45  \\ %160a@0.2
Progressive DVSR              & 32.55  \\ %92a@0.2
Progressive DVSR + Attention  & 34.49  \\ %138a@0.2
\hline
\end{tabular}
\end{center}
\end{table}

\section{Acknowledgement}
This research is based upon work supported by the Office of the Director of National
Intelligence (ODNI), Intelligence Advanced Research Projects Activity (IARPA), via IARPA R\&D Contract No. D17PC00345. The views and conclusions contained herein are those
of the authors and should not be interpreted as necessarily representing the official
policies or endorsements, either expressed or implied, of the ODNI, IARPA, or the U.S.
Government. The U.S. Government is authorized to reproduce and distribute reprints
for Governmental purposes notwithstanding any copyright annotation thereon.

%\clearpage
% ---- Bibliography ----
%
% BibTeX users should specify bibliography style 'splncs04'.
% References will then be sorted and formatted in the correct style.
%
\bibliographystyle{IEEEtran}
\bibliography{IEEEabrv,egbib}

%\bibliographystyle{IEEEtran}
% argument is your BibTeX string definitions and bibliography database(s)
%\bibliography{IEEEabrv,../bib/paper}

\end{document}